\documentclass[twoside,11pt]{article}

\usepackage{jagi}
\usepackage{amsthm}
\usepackage{amsmath}
\usepackage{amssymb}
\usepackage[all]{xy}
\usepackage{psfrag}
\usepackage{mathrsfs}
\usepackage{booktabs}
\usepackage[ruled,vlined]{algorithm2e}

\newcommand{\field}[1]{\mathbb{#1}}
\newcommand{\power}{\mathscr{P}}
\newcommand{\fs}[1]{\mathcal{#1}}

\newcommand{\nats}{\field{N}}
\newcommand{\reals}{\field{R}}
\newcommand{\define}{:=}
\newcommand{\comment}[1]{}

\newcommand{\g}[1]{\underline{#1}}
\newcommand{\defterm}[1]{\textbf{#1}}

\newcommand{\drawnfrom}{\sim}       % Drawn from
\newcommand{\gend}{\mathbf{G}}      % Generative Probability
\newcommand{\refd}{\mathbf{R}}      % Reference Probability
\newcommand{\prob}{\mathbf{P\!r}}   % Probability
\newcommand{\probc}{{P\!r}}         % Probability causal
\newcommand{\agent}{\mathbf{P}}     % Agent
\newcommand{\agentc}{P}             % Agent causal model
\newcommand{\ragent}{\mathbf{P}_0}  % Reference Agent
\newcommand{\ragentc}{P_0}          % Reference Agent causal model
\newcommand{\env}{\mathbf{Q}}       % Environment Probability
\newcommand{\envc}{Q}               % Environment Probability
\newcommand{\obs}{\text{obs}}       % Observable
\newcommand{\utility}{\mathbf{U}}   % Utility
\newcommand{\ext}[1]{#1} % External
\newcommand{\info}{\mathbf{h}}      % Information content
\newcommand{\energy}{\mathbf{e}}    % Energy
\newcommand{\fe}{\mathbf{F}}        % Free energy
\newcommand{\fu}{\mathbf{J}}        % Free utility

\newcommand{\expect}{\mathbf{E}}    % Expectation
\theoremstyle{plain}
\newtheorem{theorem}{Theorem}

\theoremstyle{definition}
\newtheorem{definition}[theorem]{Definition}

\theoremstyle{remark}

\jagiheading{x}{2010}{x}{x/xx}{xx/xx}{Pedro A. Ortega and Daniel A. Braun}%

\ShortHeadings{An axiomatic formalization of bounded rationality}{Ortega \& Braun} \firstpageno{1}

\title{An axiomatic formalization of bounded rationality \\ %
       based on a utility-information equivalence}

\author{\name Pedro A. Ortega \email peortega@dcc.uchile.cl \\
       \addr Department of Engineering\\
       University of Cambridge\\
       Cambridge CB2 1PZ, UK
       \AND
       \name Daniel A. Braun \email dab54@cam.ac.uk \\
       \addr Department of Engineering\\
       University of Cambridge\\
       Cambridge CB2 1PZ, UK}

\editor{X}

\begin{document}

\maketitle

\begin{abstract}%
Classic decision-theory is based on the maximum expected utility (MEU) principle, but crucially ignores the resource costs incurred when determining optimal decisions. Here we propose an axiomatic framework for bounded decision-making that considers resource costs. Agents are formalized as probability measures over input-output streams. We postulate that any such probability measure can be assigned a corresponding conjugate utility function based on three axioms: utilities should be real-valued, additive and monotonic mappings of probabilities. We show that these axioms enforce a unique conversion law between utility and probability (and thereby, information). Moreover, we show that this relation can be characterized as a variational principle: given a utility function, its conjugate probability measure maximizes a free utility functional. Transformations of probability measures can then be formalized as a change in free utility due to the addition of new constraints expressed by a target utility function. Accordingly, one obtains a criterion to choose a probability measure that trades off the maximization of a target utility function and the cost of the deviation from a reference distribution. We show that optimal control, adaptive estimation and adaptive control problems can be solved this way in a resource-efficient way. When resource costs are ignored, the MEU principle is recovered. Our formalization might thus provide a principled approach to bounded rationality that establishes a close link to information theory.
\end{abstract}%

\section{Introduction}
Rational decision-making is based on the principle of \emph{(subjective)
maximum expected utility} (MEU) \citep{Neumann1944, Savage1954, Anscombe1963}.
According to the MEU principle, a rational agent chooses its action $a$ so as
to maximize its expected utility
\[
    \expect[\utility|a] = \sum_s \prob(s|a) \utility(s)
\]
given the probability $\prob(s|a)$ that action $a \in \fs{A}$ will lead to
outcome $s \in \fs{S}$ and given that the desirability of the outcome $s$ is
measured by the utility $\utility(s) \in \reals$. Thus, expected utilities
express betting preferences over lotteries with uncertain outcomes. The optimal
action $a^\ast \in \fs{A}$ is defined as the one that maximizes the expected
utility, that is
\[
    a^\ast \define \arg \max_a \expect[\utility|a].
\]
What is not apparent from this simple formula, however, is that finding the
optimal action can be very difficult, especially for decision-making problems
in uncertain environments with very large space of outcomes $\fs{S}$. One could
easily imagine that computing the optimal answer is so costly (in terms of
computational resources), that one would rather content oneself with a slightly
``sub-optimal'' solution that incurs into less resource costs. The problem is,
however, that the MEU principle as stated above does not formally consider resource
costs, and hence the problem of limited resources is ignored.
Attempts to take resource costs into account for efficient decision-making
have led to the important concept of (resource-)bounded rationality \citep{Simon1982}.

In this paper we propose an axiomatic formalization of bounded rationality
that interprets a decision-maker's behavior (characterized by a probability
measure) as an implicit manifestation of his preferences. We postulate three
axioms that lead to a quantitative conversion between utilities and
probabilities (and ultimately, information), which establishes a
duality between the probability- and utility-representation of a
decision-maker. We show that the link between these representations can be
characterized by a variational principle, which allows interpreting the
probability measure as the equilibrium distribution over a constraint landscape
determined by the utility function. Based on this interpretation, we then formalize
the problem of maximizing the expectation of a target utility function as a
transformation of an initial probability measure (encoding the prior behavior
of the decision-maker) into a final probability measure that considers
both the deviation from the initial probability measure and the new constraint
given by the target utility function. We
show how this leads to a principled way to choose a probability measure that
optimally trades off the benefits of maximizing the target utilities against
the costs of transforming the probability measure. We apply this formalism
to stochastic systems that process an input-output (I/O) stream
in a sequential fashion and construct a generalized variational principle
for this setup. Finally, we show how to apply this generalized
principle to derive solutions to the problems of optimal control,
adaptive estimation and adaptive control.

\section{Conversion between probability and utility}

\subsection{Preliminaries and notation}\label{sec:preliminaries}

We introduce the following notation. A \defterm{set} is denoted by a
calligraphic letter like $\fs{X}$ and consists of \defterm{elements} or
\defterm{symbols}. \defterm{Strings} are finite concatenations of symbols.
The \defterm{empty string} is denoted by $\epsilon$. $\fs{X}^n$ denotes the set
of strings of length $n$ based on $\fs{X}$. For substrings, the following
shorthand notation is used: a string that runs from index $i$ to $k$ is written
as $x_{i:k} \define x_i x_{i+1} \ldots x_{k-1} x_k$. Similarly, $x_{\leq i}
\define x_1 x_2 \ldots x_i$ is a string starting from the first
index. By convention, $x_{i:j} \define \epsilon$ if $i>j$. Logarithms are
always taken with respect to base 2, thus $\log(2) = 1$. The symbol
$\power(\fs{X})$ denotes the powerset of $\fs{X}$, i.e.\ the set of all subsets
of $\fs{X}$.

To simplify the exposition, all probability spaces are assumed to be finite.
Due to this, we clarify some terminology. A \defterm{probability space} is a triple $(\Omega, \fs{F}, \agent)$ where $\Omega$ is
the sample space, $\fs{F} \define \power(\Omega)$ is the $\sigma$-algebra of
events, and $\agent$ is the probability measure over $\fs{F}$. A
\defterm{sample} or \defterm{outcome} is an element $\omega \in \Omega$. An
\defterm{event} is a member of $\fs{F}$ and hence a finite set of outcomes. An
\defterm{atom} is a singleton $\{\omega\} \in \fs{F}$. A \defterm{random variable} is a function $X: \Omega \rightarrow \fs{X}$ mapping each outcome $\omega$ into a symbol $X(\omega)$ from a finite alphabet $\fs{X}$. The probability of the random variable $X$ taking on the value $x \in \fs{X}$ is defined as $\agent(x) \define \agent(X=x) \define \agent(\{ \omega \in \Omega : X(\omega) = x \})$.

\subsection{Utility}\label{sec:utility}

Consider a stochastic system whose behavior is represented by a probability space $(\Omega, \fs{F}, \agent)$. The probability measure $\agent$ fully characterizes the generative law of the \emph{potential} events that the system can obtain. Thus, if $\agent(A) > \agent(B)$, then the propensity of $A$ is higher than that of $B$. This difference in probability can be given a \emph{teleological} interpretation: \emph{$A$ is more probable than $B$ because $A$ is more desirable than $B$.} For reasons that will become apparent, a measure that quantifies such differences in desirability is called a \defterm{utility} function. If there is such a measure, then it is reasonable to demand the following three properties:
\begin{itemize}
    \item[i.] Utilities should be mappings from conditional events into real numbers.
    \item[ii.] Utilities should be additive up to an arbitrary translation constant\footnote{
    That is, the utility of a joint event should be obtained by summing up the
    utilities of the sub-events (up to an arbitrary translation constant).
    The translation constant accounts for the fact that absolute values of utilities
    are not meaningful: only differences between utilities matter.
    For example,
    the ``utility of drinking coffee and eating a croissant'' should equal
    ``the utility of drinking coffee'' plus the ``utility of having a croissant
    given the reward of drinking coffee'' minus a translation constant.}.
    \item[iii.] A more probable event should have a higher utility than a less
    probable event.
\end{itemize}
The three properties can then be summarized as follows.

\begin{definition}
Let $(\Omega, \fs{F}, \agent)$ be a probability space. A function $\utility$ is
a \defterm{utility function} for $\agent$ iff it has the following three
properties for all events $A, B, C, D \in \fs{F}$ and some constant $\beta \in
\reals$:
\begin{align*}
    \text{i.}\quad
        & \utility(A|B) \in \reals,
        && \text{(real-valued)}\\
    \text{ii.}\quad
        & \utility(A \cap B|C) = \utility(A|C) + \utility(B|A \cap C) - \beta,
        && \text{(additive)}\\
    \text{iii.}\quad
        & \agent(A|B) > \agent(C|D)
            \quad \Leftrightarrow \quad
            \utility(A|B) > \utility(C|D).
        && \text{(monotonic)}
\end{align*}
\end{definition}

Furthermore, we use the abbreviation $\utility(A) \define \utility(A|\Omega)$
for ``unconditional'' events. From property (ii) it is seen that the
translation $\utility'(\cdot) = \utility(\cdot) - \beta$ leads to a strict
additivity of $\utility'$:
\begin{align*}
    \utility(A \cap B) &= \utility(A) + \utility(B|A) - \beta, \\
    (\utility'(A \cap B) + \beta) &=
        (\utility'(A) + \beta) + (\utility'(B|A) + \beta) - \beta, \\
    \utility'(A \cap B) &=
        \utility'(A) + \utility'(B|A). \\
\end{align*}
The following theorem shows that these three properties enforce a strict
mapping between probabilities and utilities.

\begin{theorem}\label{theo:utility}
If $f$ is such that $\utility(A|B) =
f(\agent(A|B))$ for any probability space $(\Omega, \fs{F}, \agent)$, then $f$
is of the form
\[
    f(\cdot) = \alpha \log(\cdot) + \beta,
\]
where $\alpha > 0$ is arbitrary strictly positive constant and $\beta$ is an
arbitrary constant.
\end{theorem}
\begin{proof}
Let $f$ be such that $f(\agent(C|D)) = \utility(C|D)$ for all $C, D \in
\fs{F}$. Let $A_1, A_2, \ldots, A_n \in \fs{F}$ be a sequence of events such
that $\agent(A_1) = \agent(A_i|\bigcap_{j<i} A_j)
> 0$ for all $i = 1,\ldots,n$. Applying $f$ yields the equivalence
\[
    \agent(A_1) = \agent\Bigl(A_i\Bigl|\bigcap_{j<i} A_j\Bigr)
    \qquad \Longleftrightarrow \qquad
    \utility(A_1) = \utility\Bigl(A_i\Bigl|\bigcap_{j<i} A_j\Bigr)
\]
for all $i=1,\ldots,n$. Using the previous properties, the product rule for
probabilities and the additivity property for utilities, one can show
\begin{align*}
    f\bigl( \agent(A_1)^n \bigr)
    &= f\biggl( \prod_{i=1}^n \agent\Bigl(A_i\Bigl|\bigcap_{j<i} A_j\Bigr) \biggr)
    = f\bigl( \agent(A_1 \cap \cdots \cap A_n) \bigr)
    = \utility(A_1 \cap \cdots A_n) \\
    &= \sum_{i=1}^n \biggl( \utility\Bigl(A_i\Bigl|\bigcap_{j<i} A_j\Bigr) - \beta \biggr)
    = n\bigl( \utility(A_1) - \beta \bigr)
    = n\bigl( f(\agent(A_1)) - \beta \bigr).
\end{align*}
Since $\agent(A_1)$ is arbitrary, this means that
\[
    f(p^n) = n(f(p) - \beta)
\]
for arbitrary $p \in (0,1]$ and $n \in \nats$.

The rest of the argument parallels Shannon's entropy theorem
\citep{Shannon1948}. Let $p,q \in (0,1]$ such that $q < p$. Choose an
arbitrarily large $m \in \nats$ and find an $n \in \nats$ to satisfy $q^m \leq
p^n < q^{m+1}$. Taking the logarithm, and dividing by $n \log q$ one obtains
\begin{equation}\label{eq:sandwich-1}
    \frac{m}{n} < \frac{\log p}{\log q} < \frac{m}{n} + \frac{1}{n}.
\end{equation}
Similarly, using $f(p^n) = n (f(p)-\beta)$ and the monotonicity of $f$, we have
\[
\begin{array}{cccccc}
    &
    q^m &<& p^n &<& q^{m+1} \\
    \Longleftrightarrow \qquad &
    f(q^m) &<& f(p^n) &<& f(q^{m+1}) \\
    \Longleftrightarrow \qquad &
    m (f(q)-\beta) &<& n (f(p)-\beta) &<& (m+1)(f(q)-\beta).
\end{array}
\]
Dividing the last set of inequalities by $n (f(p)-\beta)$ yields
\begin{equation}\label{eq:sandwich-2}
    \frac{m}{n} < \frac{f(p)-\beta}{f(q)-\beta} < \frac{m}{n} + \frac{1}{n}.
\end{equation}
Combining the inequalities in~(\ref{eq:sandwich-1}) and~(\ref{eq:sandwich-2}),
one gets
\[
    \Bigl| \frac{\log p}{\log q} - \frac{f(p)-\beta}{f(q)-\beta} \Bigr| <
    \frac{2}{n}.
\]
Since $m, n$ can be chosen arbitrary large, this implies
\[
    \frac{\log p}{\log q} = \frac{f(p)-\beta}{f(q)-\beta}
\]
in the limit $n \rightarrow \infty$. Fixing $q$ and rearranging terms gives the
functional form
\[
    f(p) = \alpha \log  p + \beta,
\]
where $\alpha$ must be positive to satisfy the monotonicity property.
\end{proof}

Thus, Theorem~\ref{theo:utility} establishes the relation
\[
    \utility(A|B) = \alpha \log \agent(A|B) + \beta,
\]
and in particular,
\[
    \utility(\Omega) = \beta.
\]
In general, if a probability measure $\agent$ and a utility function $\utility$
satisfy this relation, then we say that they are \defterm{conjugate}.
Given that this transformation is a bijection, one has that
\[
    \agent(A|B)
    = \exp\bigl\{ \tfrac{1}{\alpha} (\utility(A|B) - \utility(\Omega)) \bigl\}.
\]
There are two important observations with respect to this particular functional
form. First, note that $\info(A|B) \define -\log \agent(A|B)$ is just the Shannon
information content of $A$ given $B$. Therefore,
\[
    \utility(A|B) = -\alpha \info(A|B) + \beta.
\]
Second, this transformation implies that the probability measure $\agent$ is
the Gibbs measure with temperature $\alpha$ and energy levels $\energy(\omega)
\define -\utility(\{\omega\})$, i.e.\ the measure given by
\[
    \agent(A) =
    \frac{ \sum_{\omega \in A}
            \exp(-\frac{1}{\alpha} \energy(\omega)) }
         { \sum_{\omega \in \Omega}
            \exp(-\frac{1}{\alpha} \energy(\omega)) }
\]
for all $A \in \fs{F}$. In statistical mechanics, the Gibbs measure is the
equilibrium distribution for a given energy landscape. For this reason, we call
$\alpha > 0$ the \defterm{temperature}. The definition of utility extends to
random variables in the natural way. Thus, given a random variable $X$ with
values in $\fs{X}$, the utility of $x \in \fs{X}$ is given by $\utility(x) =
\alpha \log \agent(x) + \beta$.

\subsection{Variational principle}\label{sec:variational-principle}

The conversion between probability and utility established in
Theorem~\ref{theo:utility} satisfies a variational principle.

\begin{theorem}
Let $X$ be a random variable with values in $\fs{X}$. Let $\agent$ and
$\utility$ be a conjugate pair of probability measure and utility function over
$X$. Define the \defterm{free utility} functional as
\[
    \fu(\prob; \utility) \define \sum_{x \in \fs{X}} \prob(x) \utility(x)
        - \alpha \sum_{x \in \fs{X}} \prob(x) \log \prob(x),
\]
where $\prob$ is an arbitrary probability measure over $X$. Then,
\[
    \fu(\prob; \utility) \leq \fu(\agent; \utility) = \utility(\Omega).
\]
\end{theorem}
\begin{proof}
A similar proof to the present one is given in \citet[Theorem
1.1.3]{Keller1998}. Rewriting terms using the utility-probability conversion
and applying Jensen's inequality yields
\begin{align*}
    \fu(\prob; \utility)
    &= \sum_{x \in \fs{X}} \prob(x) \utility(x)
        - \alpha \sum_{x \in \fs{X}} \prob(x) \log \prob(x) \\
    &= \alpha \sum_{x \in \fs{X}} \prob(x)
        \log \frac{\exp(\tfrac{1}{\alpha} \utility(x))}{\prob(x)} \\
    &\leq \alpha \log \sum_{x \in \fs{X}} \prob(x)
        \frac{\exp(\tfrac{1}{\alpha} \utility(x))}{\prob(x)} \\
    &= \alpha \log \sum_{x \in \fs{X}} \agent(x) \exp(\utility(\Omega)) \\
    &= \utility(\Omega),
\end{align*}
with equality iff $\frac{\exp(\tfrac{1}{\alpha} \utility(x))}{\prob(x)}$ is
constant, i.e.\ if $\prob = \agent$.
\end{proof}

The free utility\footnote{The functional $\fe
\define -\fu$ is also known as the \emph{Helmholtz free energy} in
thermodynamics. $\fe$ is a measure of the ``useful'' work obtainable from a
closed thermodynamic system at a constant temperature and volume.} is the
expected utility of the system plus the uncertainty over the outcome. The
variational principle tells us that the probability law $\agent$ of the system
is the one that maximizes the free utility for a given utility function
$\utility$, since
\[
    \agent = \arg \max_\prob \fu(\prob; \utility).
\]
Here the utility
function $\utility$ plays the role of a constraint landscape for the
probability measure $\agent$. As the temperature $\alpha$ approaches zero, the
probability measure $\agent(x)$ approaches a delta function
$\delta_{x^\ast}(x)$, where $x^\ast = \arg \max_x \utility(x)$. Similarly, as
$\alpha \rightarrow \infty$, $\agent(x) \rightarrow \frac{1}{|\fs{X}|}$, i.e.\
the uniform distribution over $\fs{X}$. Hence, the temperature $\alpha$ plays
the role of the conversion factor between resources and utilities.

\begin{figure}[htbp]
\centering %
\psfrag{S1}[c]{$\agent_i, \utility_i$} %
\psfrag{S2}[c]{$\agent_f, \utility_f$} %
\psfrag{W}[c]{$+ \utility_*$} %
\includegraphics{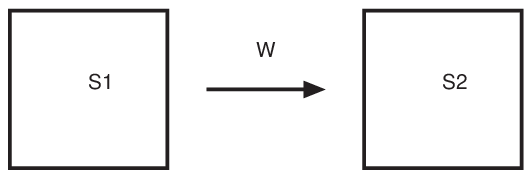}
\caption{A transformation from a system $(\agent_i, \utility_i)$ into a system
$(\agent_f, \utility_f)$ by addition of a constraint $\utility_*$.}
\label{fig:transformation} %
\end{figure}

The variational principle allows conceptualizing transformations of stochastic
systems (Figure~\ref{fig:transformation}). Consider an initial system having
probability measure $\agent_i$ and utility function $\utility_i$. This system
satisfies the equation
\[
    \fu_i \define
    \sum_{x \in \fs{X}} \agent_i(x) \utility_i(x)
    - \alpha \sum_{x \in \fs{X}} \agent_i(x) \log \agent_i(x)
    = \utility_i(\Omega).
\]
We add new constraints represented by the utility function $\utility_*$. Then,
the resulting utility function $\utility_f$ is given by the sum
\[
    \utility_f = \utility_i + \utility_*,
\]
and the resulting probability measure $\agent_f$ maximizes
\begin{align*}
    \fu(\prob, \utility_f)
    &= \sum_{x \in \fs{X}} \prob(x) \utility_f(x)
        - \alpha \sum_{x \in \fs{X}} \prob(x) \log \prob(x) \\
    &= \sum_{x \in \fs{X}} \prob(x) (\utility_i(x) + \utility_*(x))
        - \alpha \sum_{x \in \fs{X}} \prob(x) \log \prob(x) \\
    &= \sum_{x \in \fs{X}} \prob(x) \utility_*(x)
        - \alpha \sum_{x \in \fs{X}} \prob(x) \log \frac{ \prob(x) }{ \agent_i(x) }
        + \utility_i(\Omega).
\end{align*}
Let $\fu_f \define \fu(\agent_f, \utility_f)$. The difference in free utility
is
\begin{equation}\label{eq:free-utility-change}
    \fu_f - \fu_i
    = \sum_{x \in \fs{X}} \agent_f(x) \utility_*(x)
        - \alpha \sum_{x \in \fs{X}} \agent_f(x) \log \frac{ \agent_f(x) }{ \agent_i(x) }.
\end{equation}
The difference in free utility has an interpretation that is crucial for the formalization of bounded rationality: it is the expected target utility $\utility_*$ (first term) penalized by the cost of transforming $\agent_i$ into $\agent_f$ (second term). Clearly, (\ref{eq:free-utility-change}) is a functional to be maximized. Depending on the givens and the unknowns, this leads to different variational problems. We emphasize the two cases that are important for our exposition:
\begin{enumerate}
    \item \emph{Control.} If we fix the initial probability measure $\agent_i$ and
    the constraint utilities $\utility_*$, then the final system $\agent_f$ optimizes
    the trade-off between utility and resource costs. That is,
    \begin{equation}\label{eq:fu-endogenous}
    \agent_f = \arg \max_\prob
    \sum_{x \in \fs{X}} \prob(x) \utility_*(x)
        - \alpha \sum_{x \in \fs{X}} \prob(x) \log \frac{ \prob(x) }{ \agent_i(x) }.
    \end{equation}
    The solution is given by
    \[
        \agent_f(x) \propto \agent_i(x)
            \exp\biggl( \frac{1}{\alpha} \utility_*(x) \biggr).
    \]
    In particular, at very low temperature $\alpha \approx 0$,
    (\ref{eq:free-utility-change}) becomes
    \[
        \fu_f - \fu_i
        \approx \sum_{x \in \fs{X}} \agent_f(x) \utility_*(x),
    \]
    and hence resource costs are ignored in the choice of $\agent_f$, leading to
    $\agent_f \approx \delta_{x^\ast}(x)$, where $x^\ast = \max_x \utility_*(x)$. Similarly, at
    a high temperature, the difference is
    \[
        \fu_f - \fu_i
        \approx - \alpha \sum_{x \in \fs{X}} \agent_f(x)
            \log \frac{ \agent_f(x) }{ \agent_i(x) },
    \]
    and hence only resource costs matter, leading to $\agent_f \approx \agent_i$.
    \item \emph{Estimation.} If we fix the final probability measure $\agent_f$
    and the constraint utilities $\utility_*$, then the initial system $\agent_i$
    satisfies
    \begin{align}
        \agent_i &= \arg \max_\prob
        \sum_{x \in \fs{X}} \agent_f(x) \utility_*(x)
            - \alpha \sum_{x \in \fs{X}} \agent_f(x) \log \frac{ \agent_f(x) }{ \prob(x) }
            \label{eq:fu-exogenous} \\
        &= \arg \min_\prob
        \sum_{x \in \fs{X}} \agent_f(x) \log \frac{ \agent_f(x) }{ \prob(x) },
            \nonumber
    \end{align}
    and thus we have recovered the minimum relative entropy principle for
    estimation, having the solution
    \[
        \agent_i = \agent_f.
    \]
    Varying the initial distribution $\agent_i$ is equivalent to varying the utility $\utility_i$
    as part of $\utility_f$ such that the given distribution $\agent_f$ becomes the equilibrium distribution.
\end{enumerate}
Alternatively, one can regard \emph{control} as the problem of finding $\agent_f$ given $\utility_*$ and $\utility_i$; and \emph{estimation} as the problem of finding $\utility_i$ given $\agent_f$ and $\utility_*$. This is easily seen after rewriting the terms in~(\ref{eq:free-utility-change}).

\section{I/O systems}\label{sec:io-system}

We now turn our discussion to I/O systems. Informally, I/O systems model anything that has an \emph{I/O stream}, like a calculator, a human cell, an animal, a computer program or a robot. In this sense, an I/O system is not required to be a discretely identifiable (physical) entity as long as there is a viewpoint from which it \emph{appears} to have an I/O stream. For example, from a robot's perspective, its \emph{environment} is a well-defined system too because it has an ``input channel'' to absorb the robot's actions and an ``output channel'' to produce the robot's perceptions.

The mathematical description of an I/O system can be done at several levels. This paper focusses on two of them: behavior and beliefs. A \emph{model of behavior} is a direct specification of an I/O system that merely describes the statistics of the I/O stream. A \emph{model of beliefs} is an indirect specification of an I/O system that has the advantage of representing the I/O system's underlying assumptions that give rise to its behavior.

\subsection{Model of behavior}

\begin{figure}[htbp]
\centering %
\psfrag{S1}[c]{$\agent$} %
\psfrag{S2}[c]{$\env$} %
\psfrag{X1}[c]{$X_1$} %
\psfrag{X2}[c]{$X_2$} %
\psfrag{X3}[c]{$X_3$} %
\psfrag{XT}[c]{$X_T$} %
\includegraphics{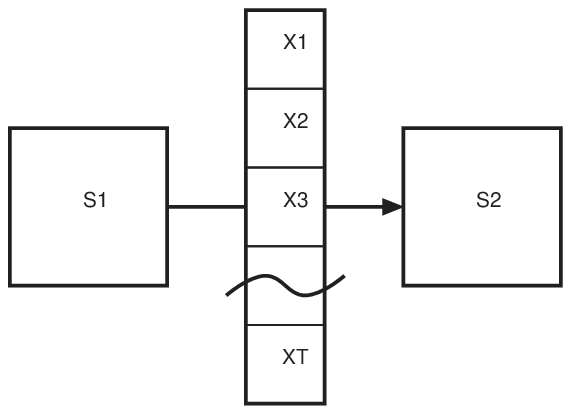}
\caption{Two I/O systems $\agent$ and $\env$ interacting with each other.}
\label{fig:io-system} %
\end{figure}

Formally, an I/O system is an abstract model of a (stochastic) machine that processes input symbols and generates output symbols. These symbols are exchanged with another (external) I/O system via an I/O channel (Figure~\ref{fig:io-system}).

The interaction between two I/O systems proceeds in cycles $t=1, 2, \ldots, T$ following a predefined protocol. The protocol determines which system is responsible for each cycle. In cycle $t$, the responsible system generates a symbol $x_t$ conditioned on the past symbols $x_{<t}$. Then the cycle $t+1$ starts.

If one wants to characterize the way an I/O system \emph{behaves}, it is necessary to specify the statistics governing its potential I/O stream. One can encapsulate all the details by providing the probability distribution over the potential I/O sequences.

\begin{definition}
An \defterm{I/O system} is a probability measure $\agent$ over $T$ random variables $X_1, X_2, \ldots, X_T$ taking on values in finite alphabets $\fs{X}_1, \fs{X}_2, \ldots, \fs{X}_T$.
\end{definition}

Because the I/O system processes both input and output symbols, the probability measure $\agent$ contains both evidential and generative probabilities. The evidential probabilities, called \defterm{plausibilities}, allow the I/O system to infer properties about its input stream; while the generative probabilities, called \defterm{propensities}, prescribe the law to generate its output stream. Hence, if $x_t$ is generated by an external I/O system, then $\agent(x_t|x_{<t})$ is the plausibility of observing $x_t$ given the past I/O string $x_{<t}$; while if $x_t$ is generated by the I/O system $\agent$ itself, then $\agent(x_t|x_{<t})$ is the propensity of producing $x_t$ given the past I/O string $x_{<t}$.

\subsection{Model of beliefs}

While the previous definition contains all the necessary details to describe the behavior of an I/O system, it falls short modeling the I/O system's underlying assumptions that bring about its behavior. Importantly, it is desirable to model the uncertainties two interacting I/O systems have about each other, because these uncertainties play a fundamental r\^{o}le in conceptualizing adaptive behavior. The aim of this section is to introduce a model for I/O systems that allows explicitly representing these uncertainties.

\subsubsection{Causal Models}

From the point of view of an I/O system~$\agent$ that is interacting with an I/O system~$\env$, one needs to represent (a)~the uncertainty~$\agent$ has about~$\env$ and~(b) the uncertainty~$\env$ has about~$\agent$. Following a Bayesian approach, both uncertainties are modeled by the introduction of hidden/undisclosed variables. More specifically, cases~(a) and~(b) can be modeled by undisclosed inputs and undisclosed outputs respectively, i.e.\ symbols that are generated but kept hidden from the other system\footnote{\emph{Undisclosed inputs}, commonly known as \emph{hypotheses} or \emph{latent variables} in Bayesian statistics, are at the heart of Bayesian inference \citep{Jaynes2003}. In game theory, \emph{undisclosed outputs} determine the \emph{player types}. Player types are the crucial component of a Bayesian game whose purpose is to model games with incomplete information \citep{Osborne1999}.}. The inclusion of undisclosed random variables requires extending the interaction model as follows.

The interaction between two I/O systems proceeds in cycles $t=1, 2, \ldots, T$. In each cycle, either one of the two systems generates a symbol $x_t$ conditioned on the previously observed symbols. The symbol $x_t$ might be either disclosed or undisclosed. A disclosed symbol is observed by both systems, while an undisclosed one is only observed by the system who generated it. After a symbol is generated, the I/O systems that have observed it update their belief states.

To illustrate how uncertainty is modeled, consider the familiar Bayesian estimator. Let $\fs{D} \define \fs{D}_1 \times \ldots \times \fs{D}_N$ be a set of strings, where each $\fs{D}_n$, $1 \leq n \leq N$, is a finite alphabet. A Bayesian estimator over $\fs{D}$ with hypotheses $\Theta$ is a probability measure $P$ over $\Theta \times \fs{D}$ of the form
\begin{equation}\label{eq:bayesian-estimator}
    \agentc(d_{\leq N})
    = \sum_{\theta \in \Theta} \agentc(d_{\leq N}|\theta) \agentc(\theta),
\end{equation}
where: $d_{\leq N}$ is an observation string with $d_n \in \fs{D}_n$ for all $1 \leq n \leq N$; $\theta \in \Theta$ is a hypothesis; $P(d_{\leq N}|\theta)$ is the likelihood of $d_{\leq N}$ under the hypothesis $\theta$; and $P(\theta)$ is the prior probability of the hypothesis $\theta$. The Bayesian estimator is an adaptive predictor: it uses the symbols observed in the past to predict the next symbol. The predictive distribution over the $n$-th observation ($1 \leq n \leq N$) conditioned on the past observations $d_{<n}$ is then given by
\begin{equation}\label{eq:predictive-distribution}
    \agentc(d_n|d_{<n})
    = \sum_{\theta \in \Theta} \agentc(d_n|\theta, d_{<n})
        \agentc(\theta|d_{<n}),
\end{equation}
where $\agentc(d_n|\theta, d_{<n})$ is the likelihood of $d_n$ under hypothesis $\theta$ given the past observations~$d_{<n}$ and $\agentc(\theta|d_{<n})$ is the posterior probability of $\theta$ given the past observations $d_{<n}$. Both of these quantities are obtained from $\agentc(\theta, d_{\leq N})$ by applying standard probability calculus. It is easy to see that this probabilistic model corresponds to an I/O system $\agent$ over a sequence $x_{\leq T}$ where: $T \define N+1$; $x_1 \define \theta$ is an undisclosed input drawn from $\fs{X}_1 \define \Theta$; and $x_t \define d_{t-1}$ ($2 \leq t \leq T$) is a disclosed input drawn from $\fs{X}_t \define \fs{D}_{t-1}$. The probability measure $\agent$ is constructed from $P$ as
\[
    \agent(\theta, d_{\leq N}) \define
    \agentc(\theta) \agentc(d_1) \agentc(d_2|d_1)
        \cdots \agentc(d_N|d_{<N}),
\]
where one has to notice that $\agent \neq \agentc$ because $\theta$ is unobserved and thus cannot be used to condition , i.e.\
\[
    \agent(d_n|\theta, d_{<n})
    = \agentc(d_n|d_{<n})
    = \sum_{\theta' \in \Theta} \agentc(d_n|\theta', d_{<n}) \agentc(\theta')
    \neq \agentc(d_n|\theta, d_{<n}).
\]
Hence, this illustrates two facts. First, undisclosed inputs play the role of hypotheses. Second, the model $\agentc$ and the I/O system $\agent$ are in general not the same.

Extending this scheme to include outputs as well is not straightforward. If some of the~$d_n$ are generated by the system itself, then~(\ref{eq:predictive-distribution}) does not hold anymore, because outputs are syntactically different from inputs, requiring belief updates governed by causal constraints. Essentially, an input provides the system with information about the whole history of the stochastic process, while an output, by virtue of being generated \emph{by the system itself as a function of the past}, provides the system only with information about the present and future of the stochastic process because the past cannot be changed. See for instance \citet{Shafer1996}, \citet{Pearl2000}, \citet{Spirtes2001} and \citet{Dawid2010} for a more in-depth exposition of causality.

In order carry out the belief updates following outputs, it is necessary to know the \emph{causal probability model} for $P$. The causal probability model consists of a set of conditional probability measures highlighting the  functional dependencies amongst the random variables. This is reflected in the following definition.

\begin{figure}[htbp]
\centering %
\psfrag{L1}[c]{output} %
\psfrag{L2}[c]{input} %
\psfrag{L3}[c]{disclosed} %
\psfrag{L4}[c]{undisclosed} %
\psfrag{L5}[c]{observable} %
\psfrag{S1}[c]{$\agentc$} %
\psfrag{S2}[c]{$\envc$} %
\psfrag{S3}[c]{$\agentc$} %
\psfrag{S4}[c]{$\envc$} %
\psfrag{S5}[c]{$\agentc$} %
\psfrag{S6}[c]{$\envc$} %
\psfrag{S7}[c]{$\agentc$} %
\psfrag{S8}[c]{$\envc$} %
\psfrag{X1}[c]{$X$} %
\psfrag{X2}[c]{$X$} %
\psfrag{X3}[c]{$X$} %
\psfrag{X4}[c]{$X$} %
\includegraphics{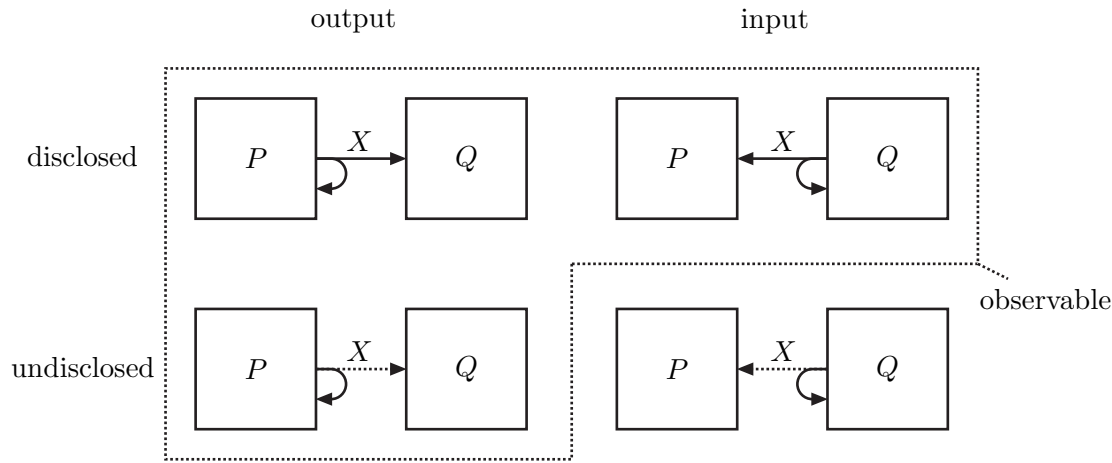}
\caption{The four types of random variables with respect to $\agentc$. Solid
arrows mean that the value of the random variable is disclosed, while dashed arrows mean that the value is undisclosed. The enclosed area contains the random variables that are observable by $\agentc$.}
\label{fig:endo-exo} %
\end{figure}

\begin{definition}
A \defterm{causal model} of an I/O system is a set of $T$ conditional probability measures $\agentc(X_1), \agentc(X_2|X_1), \ldots, \agentc(X_T|X_{<T})$ over \emph{typed} random variables $X_1, X_2, \ldots, X_T$ taking on values in finite alphabets $\fs{X}_1, \fs{X}_2, \ldots, \fs{X}_T$ .
\end{definition}

The causal model explains how the random variables functionally depend on each other. In particular, for all $t \geq 1$, the value of $X_t$ is generated \emph{as a function} of the values of $X_1, \ldots, X_{t-1}$. The probability measure $P$ over all the random variables is obtained by the product rule:
\[
    \agentc(X_1,\ldots,X_T) \define \prod_{t=1}^T \agentc(X_t|X_{<t}).
\]
For notational convenience, we will use the letter $P$ as a shorthand for the whole causal model.

The \defterm{type} of a random variable specifies whether it an input or an output, and whether it is disclosed or undisclosed (Figure~\ref{fig:endo-exo}). Both distinctions give rise to $2 \times 2 = 4$ possible types. If a random variable $X_t$ is \emph{not} an undisclosed input, then we say that it is \defterm{observable}. In this sense, being or not observable is not a type, but a property of the random variable. The operational significance of the type of random variables will become clear in the context of belief updates.

\subsubsection{Belief updates}

\begin{figure}[htbp]
\centering %
\psfrag{t1}[c]{$X_2 = 1$} %
\psfrag{t2}[c]{$X_2 \leftarrow 1$} %
\psfrag{x1}[c]{$X_1$} %
\psfrag{x2}[c]{$X_2$} %
\psfrag{x3}[c]{$X_3$} %
\psfrag{x4}[c]{$X_1$} %
\psfrag{x5}[c]{$X_2$} %
\psfrag{x6}[c]{$X_3$} %
\psfrag{x7}[c]{$X_1$} %
\psfrag{x8}[c]{$X_2$} %
\psfrag{x9}[c]{$X_3$} %
\psfrag{xa}[c]{$X_1$} %
\psfrag{xb}[c]{$X_2$} %
\psfrag{xc}[c]{$X_3$} %
\psfrag{xd}[c]{$X_1$} %
\psfrag{xe}[c]{$X_2$} %
\psfrag{xf}[c]{$X_3$} %
\includegraphics{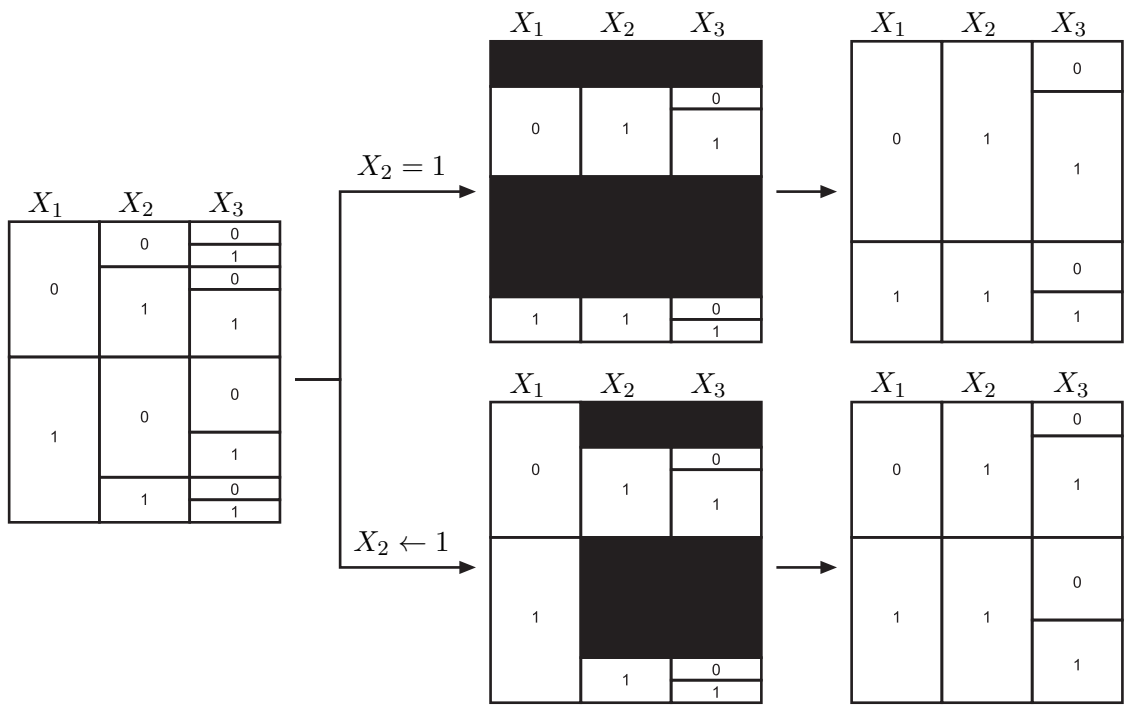}
\caption{A logical versus a causal update. The figure shows three causally
ordered random variables $X_1$, $X_2$ and $X_3$ (taking on binary values) and
their probabilities (through the height of their boxes). Two updates are
compared: the logical update $X_2 = 1$ and the causal update $X_2 \leftarrow
1$. These updates eliminate the incompatible probability mass (as shown in the
first column after the update) and then normalize the remaining probability
mass (second column after the update). Note that a logical update affects the
probability mass of the whole history, eliminating the incompatible
realizations; while a causal update affects only the probability mass of the
present and the future.}
\label{fig:logical-causal} %
\end{figure}

When an I/O system observes the value $x_t \in \fs{X}_t$ of a random variable $X_t$, then its information state is updated. This update depends on whether $X_t$ is an input or an output. If $X_t$ is an input, then the update is \defterm{logical}. If $X_t$ is an output, then the update is \defterm{causal}. This difference is illustrated in Figure~\ref{fig:logical-causal}.

A logical update models a \defterm{measurement}. As such, it provides information about the whole realization of the stochastic process. That is, learning the value of $X_t$ provides information about all $\{ X_s : t \leq s \leq T \}$ through the dependencies established by the causal model for $P$. A logical update $X_t = x_t$ changes all conditional probabilities as
\[
    \agentc(A|B) \quad \xrightarrow{X_t = x_t}
    \quad \agentc(A|B, X_t=x_t),
\]
where $A$ and $B$ are arbitrary events. The plausibility of observing a sequence $x_1, x_2, \ldots, x_t$ (in this order) is given by
\[
    \agentc(x_1) \agentc(x_2|x_1) \agentc(x_3|x_1,x_2)
        \cdots \agentc(x_t|x_1,\ldots,x_{t-1})
    = \agentc(x_{\leq t}),
\]
where the last equality follows from basic probability calculus.

A causal update models a \defterm{decision}. As such, it only provides information about the future of the realization of the stochastic process, but not about its past. That is, learning the value of $X_t$ provides information about $\{ X_s : s \geq t, s \in \nats \}$ \emph{only}. Furthermore, the random variable $X_t$ is rendered independent from its past, thereby reflecting the autonomy of the decision. A causal update $X_t \leftarrow x_t$ changes all conditional probabilities as
\[
    \agentc(A|B) \quad \xrightarrow{X_t \leftarrow x_t}
    \quad \agentc(A|B, X_t \leftarrow x_t) = \agentc'(A|B, X_t=x_t),
\]
where $A$ and $B$ are arbitrary events and where $\agentc'$ is the probability measure uniquely defined by the equations
\begin{equation}
\label{eq:causal-transform}
\begin{aligned}
    \text{i.}  \qquad &\agentc'(X_{<t})
        = \agentc(X_{<t}),
    &&\text{(past)}\\
    \text{ii.} \qquad &\agentc'(X_t|X_{<t})
        = \delta_{x_t}(X_t),
    &&\text{(present)}\\
    \text{iii.}\qquad &\agentc'(X_{t+1:T}|X_{\leq t})
        = \agentc(X_{t+1:T}|X_{\leq t}).
    &&\text{(future)}
\end{aligned}
\end{equation}
When the random variable $X_t$ is clear from the context, we use the
abbreviation
\[
    \agentc(A|B, \hat{x}_t) \define \agentc(A|B, X_t \leftarrow x_t).
\]
The propensity of generating a sequence $x_1, x_2, \ldots, x_t$ (in this order) is given by
\[
    \agentc(x_1) \agentc(x_2|\hat{x}_1) \agentc(x_3|\hat{x}_1,\hat{x}_2)
        \cdots \agentc(x_t|\hat{x}_1,\ldots,\hat{x}_{t-1})
        = \agentc(x_{\leq t}),
\]
where the equality is obtained by using the definition of causal updates and then applying basic probability calculus.

When an I/O system does not observe the value $x_t \in \fs{X}_t$ of a random variable $X_t$ because it is an undisclosed input, then its information state is \emph{not} updated. That is, an unobserved update $X_t = x_t$ leaves all conditional probabilities unchanged, i.e.\ 
\[
    \xymatrix{
        \agentc(A|B) \qquad \ar@{.>}[r]^{X_t = x_t} & \qquad \agentc(A|B)
    },
\]
where $A$ and $B$ are arbitrary events.

\subsubsection{Deriving behavior from beliefs}

As anticipated previously, a model $\agentc$ of an I/O system $\agent$ gives rise to a probability measure characterizing an I/O system. The probability measure $\agent$ is derived from the causal model $\agentc$ as follows:

\begin{definition}
Let $P$ be a causal model of an I/O system. The \defterm{associated I/O system} $\agent$ is the I/O system recursively defined as
\begin{align}
    \ext{\agent}(\epsilon) &\define 1, &
    \ext{\agent}(x_{\leq t}) &\define \ext{\agent}(x_{<t})
    \agentc(x_t|\obs(x_{<t})),
\end{align}
where the auxiliary function $\obs(\cdot)$ is given by
\[
    \obs(\epsilon) \define \epsilon, \qquad
    \obs(x_{\leq t}) \define
    \begin{cases}
        \obs(x_{<t}) \hat{x}_t & \text{if $X_t$ is an output,} \\
        \obs(x_{<t}) x_t       & \text{if $X_t$ is a disclosed input,} \\
        \obs(x_{<t})           & \text{if $X_t$ is an undisclosed input.} \\
    \end{cases}
\]
\end{definition}

In this definition, $\obs(x_{\leq t})$ selects the values that the I/O system
has observed at time $t+1$, flagging them as either causal or logical belief
updates.  By construction, $\agent$ has the important property that for all $x_{\leq t}$,
\[
    \agent(x_t|x_{<t}) = \agentc(x_t|\obs(x_{<t})).
\]

\subsection{The variational principle in I/O systems}\label{sec:vp-io-systems}

Let us assume that we are in possession of a reference I/O system $\ragent$ (or its causal model~$\ragentc$) encoding our current knowledge. The problem is that we wish to convert $\ragent$ into an I/O system $\agent$ maximizing a
given target utility function $\utility_\ast$. We assume further that $\ragent$,
$\agent$ and $\utility_\ast$ share their random variables including the causal
order and types. As we have argued previously, any transformation of a
probability measure incurs into costs. These costs can potentially be so high
that they jeopardize the benefits of na\"{\i}vely maximizing the expectation of
$\utility_\ast$. We therefore seek an optimality principle that allows finding
a probability measure $\agent$ that trades off the benefits against the costs
of this transformation.

In accord with Section~\ref{sec:variational-principle}, we first not that the
transformation of the reference I/O system $\ragent$ into $\agent$ due to the
addition of constraints $\utility_\ast$ can be expressed as a change in free
utility characterized by Equation~(\ref{eq:free-utility-change}).
The free utility functional for a given conjugate pair
$(\agent, \utility)$ can be expressed as follows
\[
    \fu(\ext{\agent}; \ext{\utility}) \define
    \sum_{x_{\leq T}} \ext{\agent}(x_{\leq T}) \ext{\utility}(x_{\leq T})
    - \alpha \sum_{x_{\leq T}} \ext{\agent}(x_{\leq T}) \log \ext{\agent}(x_{\leq T}).
\]
In Section~\ref{sec:variational-principle} we have also
emphasized that there are two variational problems, namely the \emph{control} and the
\emph{estimation} problem, that arise depending on the givens and the unknowns
of the variation. Naturally, this distinction carries over in the case
of probability distributions representing I/O systems.

Suppose for simplicity that $T=1$. Thus, we have to find an I/O system $\agent$ over a single
random variable $X \define X_1$ taking on values in~$\fs{X} \define \fs{X}_1$.
Again, we write down the difference in free utility, but identifying the givens
with $\ragent$ and the unknowns with $\prob$. This yields the following two
problems.
\begin{enumerate}
    \item \emph{Control.} If we are searching for a probability law $\agent$
    that fulfills the constraints given by the maximization of $\ext{\utility}_\ast$ and the
    minimization of the cost of the transformation $\ragent \rightarrow
    \agent$, then we use Equation~(\ref{eq:fu-endogenous}), i.e.\
    \[
        \agent = \arg \max_\prob
        \sum_{x \in \fs{X}} \prob(x) \utility_\ast(x)
            - \alpha \sum_{x \in \fs{X}} \prob(x) \log \frac{ \prob(x) }{ \ragent(x) }.
    \]
    \item \emph{Estimation.} If we are searching for the best estimation $\agent$
    of the probability law $\ragent$ under the constraints $\ext{\utility}_\ast$, then
    we use Equation~(\ref{eq:fu-exogenous}), i.e.\
    \[
        \agent = \arg \min_\prob
        \sum_{x \in \fs{X}} \ragent(x) \log \frac{ \ragent(x) }{ \prob(x) }.
    \]
\end{enumerate}

The same idea extends to the case where $T \geq 1$, obtaining a functional for
the difference in free utility that spans all the random variables.
A simple way to do this is again by recursively defining two auxiliary
probability measures $\gend$ and $\refd$ as
\begin{equation}
\label{eq:vp-distributions}
\begin{aligned}
    \gend(\epsilon) &\define 1, &
    \gend(x_{\leq t}) &\define
    \begin{cases}
        \gend(x_{<t}) \prob(x_t|x_{<t}) & \text{if $X_t$ is controlled,} \\
        \gend(x_{<t}) \ragent(x_t|x_{<t}) & \text{if $X_t$ is estimated;} \\
    \end{cases}
    \\
    \refd(\epsilon) &\define 1, &
    \refd(x_{\leq t}) &\define
    \begin{cases}
        \refd(x_{<t}) \ragent(x_t|x_{<t}) & \text{if $X_t$ is controlled,} \\
        \refd(x_{<t}) \prob(x_t|x_{<t}) & \text{if $X_t$ is estimated.} \\
    \end{cases}
\end{aligned}
\end{equation}
Then, it is straightforward to see that the difference in free utility is given
by
\begin{equation}\label{eq:generalized-vp}
    \agent = \arg \max_\prob \left\{
    \sum_{x_{\leq T}} \gend(x_{\leq T}) \ext{\utility}_\ast(x_{\leq T})
    - \alpha \sum_{x_{\leq T}} \gend(x_{\leq T})
        \log \frac{ \gend(x_{\leq T}) }{ \refd(x_{\leq T}) }
        \right\}.
\end{equation}

\section{Applications}

In the following, we will illustrate applications of the variational principle for I/O systems in Equation~(\ref{eq:generalized-vp}) by deriving solutions to three problems: optimal control, adaptive estimation and adaptive control.

Let $\fs{A}$ and $\fs{O}$ be two finite sets, the first being the \defterm{set
of actions} and the second being the \defterm{set of observations}.
Furthermore, let $\Theta$ be a finite set called the \defterm{set of
parameters}. The set $\fs{Z} \define \fs{A} \times \fs{O}$ is called the
\defterm{set of interactions}, and a pair $(a,o) \in \fs{Z}$ is an \defterm{interaction}.
We will underline symbols to glue them together as in $\g{ao}_{\leq t}
\define a_1 o_1 \ldots a_t o_t$ to abbreviate strings of interactions.
Let $\agent$ and $\env$ be I/O systems. By convention, we will consider
$\agent$ the system to be designed and $\env$ an external system to be
interfaced. Accordingly, we call $\agent$ the \defterm{agent}, and $\env$ the
\defterm{environment}.

Consider the following interaction protocol. Initially, $\env$ chooses
a parameter $\theta \in \Theta$ unbeknownst to $\agent$. Then, the interaction proceeds in cycles $t=1,
2, \ldots, T$. In cycle $t$, $\agent$ randomly chooses a value $a_t$ for the
random variable $A_t$ from the set of actions $\fs{A}$ conditioned on the past
I/O symbols $\g{ao}_{<t}$. $\env$ responds by choosing a value $o_t$ for the
random variable $O_t$ from the set of observations $\fs{O}$ conditioned on the
past I/O symbols $\theta\g{ao}_{<t}a_t$. Then the next cycle starts. This
interaction protocol determines a probability law over the causally ordered
random variables $\theta, A_1, O_1, \ldots, A_T, O_T$ defined as follows:
\[
    \theta \drawnfrom \env(\theta), \qquad
    a_t|\theta,\g{ao}_{<t}    \drawnfrom \agent(a_t|\g{ao}_{<t}), \qquad
    o_t|\theta,\g{ao}_{<t}a_t \drawnfrom \env(o_t|\theta, \g{ao}_{<t}a_t).
\]
Note that with respect to $\agent$, $\theta$ is a latent variable, $A_1, \ldots,
A_T$ are outputs and $O_1, \ldots, O_T$ are observable inputs.
Similarly for $\env$, $\theta, O_1, \ldots, O_T$ are
outputs, and $A_1, \ldots, A_T$ are observable inputs. This
interaction protocol, as known by $\agent$, is summarized in
Table~\ref{tab:standard-interaction-protocol}. The applications
in the following use this protocol or a simplification of it.

\begin{table} \caption{The
standard interaction protocol as seen by the agent
$\agent$.}\label{tab:standard-interaction-protocol}
\bigskip %
\centering %
\begin{tabular}{lcccccccc}
  \toprule
           & $X_1$ & $X_2$ & $X_3$ & $X_4$ & $X_5$ & $\ldots$ & $X_{2T}$ & $X_{2T+1}$ \\
  \midrule
  Name       & $\theta$ & $A_1$    & $O_1$    & $A_2$    & $O_2$    & $\ldots$ & $A_T$    & $O_T$ \\
  Alphabet   & $\Theta$ & $\fs{A}$ & $\fs{O}$ & $\fs{A}$ & $\fs{O}$ & $\ldots$ & $\fs{A}$ & $\fs{O}$ \\
  \bottomrule
\end{tabular}
\end{table}

\subsection{Optimal control}\label{sec:optimal-control}

In optimal control problems it is generally assumed that we are given a utility
function $\utility_\ast$ and that the environment is fully known, i.e.\
$\ragentc(o_t| \g{\hat{a}o}_{<t}\hat{a}_t) = \env(o_t| \g{ao}_{<t}a_t)$.
The choice of the parameter $\theta$ can be omitted. The probability measures
$\gend$ and $\refd$ are given by
\begin{align*}
    \gend(a_t| \g{ao}_{<t}) &= \probc(a_t| \g{\hat{a}o}_{<t}),
    & \gend(o_t| \g{ao}_{<t}a_t) &= \ragentc(o_t| \g{\hat{a}o}_{<t}\hat{a}_t),
    \\
    \refd(a_t| \g{ao}_{<t}) &= \ragentc(a_t| \g{\hat{a}o}_{<t}),
    & \refd(o_t| \g{ao}_{<t}a_t) &= \probc(o_t| \g{\hat{a}o}_{<t}\hat{a}_t).
\end{align*}
Hence, the variational problem to find $\agent$ is to maximize the functional
\begin{equation}\label{eq:vp-optimal-control}
    \sum_{\theta,\g{ao}_{\leq T}} \gend(\g{ao}_{\leq T})
    \Bigg[
        \ext{\utility}_\ast(\g{ao}_{\leq T})
        - \alpha \sum_{t=1}^{T} \log
        \frac{\probc(a_t|\g{ao}_{<t})}
             {\ragentc(a_t|\g{ao}_{<t})}
        - \alpha \sum_{t=1}^{T} \log
        \frac{\ragentc(o_t|\g{ao}_{<t}a_t)}
             {\probc(o_t|\g{ao}_{<t}a_t)}
    \Bigg],
\end{equation}
which results from replacing $\gend$ and $\refd$ into~(\ref{eq:generalized-vp})
and by applying the equalities
\begin{align*}
    \probc(a_t| \g{\hat{a}o}_{<t}) &= \probc(a_t| \g{ao}_{<t}),
    & \probc(o_t|\g{\hat{a}o}_{<t}\hat{a}_t) &= \probc(o_t|\g{ao}_{<t}a_t), \\
    \ragentc(a_t|\g{\hat{a}o}_{<t}) &= \ragentc(a_t| \g{ao}_{<t}),
    & \ragentc(o_t|\g{\hat{a}o}_{<t}\hat{a}_t) &= \ragentc(o_t| \g{ao}_{<t}a_t)
\end{align*}
which are easily derived using~(\ref{eq:causal-transform}) repeatedly. The
important observation is that~(\ref{eq:vp-optimal-control}) can be seen as a
concise way of expressing a collection of independent variational problems,
where this collection contains one variational problem for each random
variable. In the variational problem for the observation probabilities we can
disregard the constraint utilities and the resource cost of the action
probabilities. The $t$-th summand of the total expected reward can then be
written as
\[
    \sum_{\g{ao}_{<t} a_t} \gend(\g{ao}_{<t}a_t)
    \Bigg[ \sum_{o_t} \ragentc(o_t|\g{ao}_{<t}a_t)
        \log \frac{\probc(o_t|\g{ao}_{<t}a_t)}
                  {\ragentc(o_t|\g{ao}_{<t}a_t)}
    \Bigg].
\]
Since varying $\probc(o_t|\g{ao}_{<t}a_t)$ does not influence the summands at
times $\neq t$, the optimal solution to this minimum relative entropy problem
is trivially obtained by $\agentc(o_t|\g{ao}_{<t}a_t) =
\env(o_t|\g{ao}_{<t}a_t)$. The variational problem with respect to the action
probabilities is a little bit more intricate, since varying the first action
probability, for example, has an impact on all subsequent conditional action
probabilities. The functional~(\ref{eq:vp-optimal-control}) can be expanded
recursively, yielding

\footnotesize %
\begin{align*}
    &\sum_{a_1} \probc(a_1) \Bigg[
      \ext{\utility}_\ast(a_1)
      - \alpha \log \frac{\probc(a_1)}{\ragentc(a_1)}
      + \sum_{o_1} \agentc(o_1|a_1) \Bigg[
        \ext{\utility}_\ast(o_1|a_1) \\
        &+ \sum_{a_2} \probc(a_2|\g{ao}_1) \Bigg[
          \ext{\utility}_\ast(a_2|\g{ao}_1)
          - \alpha \log \frac{\probc(a_2|\g{ao}_1)}{\ragentc(a_2|\g{ao}_1)}
      + \sum_{o_2} \agentc(o_2|\g{ao}_1a_2) \Bigg[
        \ext{\utility}_\ast(o_2|\g{ao}_1a_2) \\
      &+ \cdots \\
      &+ \sum_{a_T} \probc(a_T|\g{ao}_{<t}) \Bigg[
        \ext{\utility}_\ast(a_T|\g{ao}_{<T})
        -\alpha \log \frac{\probc(a_T|\g{ao}_{<T})}{\ragentc(a_T|\g{ao}_{<T})}
          + \sum_{o_T} \probc(o_T|\g{ao}_{<T}a_T)
          \ext{\utility}_\ast(o_T|\g{ao}_{<T}a_T)
      \Bigg] \cdots
      \Bigg] \Bigg] \Bigg] \Bigg],
\end{align*}
\normalsize %
The innermost variational problem is of the form
\[
    \sum_{a_T} \probc(a_T|\g{ao}_{<T})
    \Bigg[ \ext{\utility}_\ast(a_t|\g{ao}_{<T})
        + \sum_{o_T} \agentc(o_T|\g{ao}_{<T}a_T)
          \ext{\utility}_\ast(o_T|\g{ao}_{<T}a_T)
        -\alpha \log \frac{\probc(a_T|\g{ao}_{<T})}{\ragentc(a_T|\g{ao}_{<T})}
           \Bigg].
\]
As discussed previously, its solution is
\[
    \agentc(a_T|\g{ao}_{<T})
    = \frac{\ragentc(a_T|\g{ao}_{<T})}{Z^\alpha(\g{ao}_{<T})}
        \exp \Bigg\{\tfrac{1}{\alpha} \ext{\utility}_\ast(a_T|\g{ao}_{<T})
            + \tfrac{1}{\alpha} \sum_{o_T} \agentc(o_T|\g{ao}_{<T}a_T)
                \ext{\utility}_\ast(o_T|\g{ao}_{<T}a_T) \Bigg \},
\]
where $Z^\alpha(\g{ao}_{<T})$ is the normalizing constant, also known as the
partition function. Similarly, the action probabilities
$\agentc(a_t|\g{ao}_{<t})$ can be obtained as
\begin{multline*}
    \agentc(a_t|\g{ao}_{<t})
    = \frac{\ragentc(a_t|\g{ao}_{<t})}{Z^{\alpha}(\g{ao}_{<t})}
    \exp\Bigg\{
        \tfrac{1}{\alpha} \ext{\utility}_\ast(a_t|\g{ao}_{<t})
        + \tfrac{1}{\alpha} \sum_{o_t} \agentc(o_t|\g{ao}_{<t}a_t)
            \ext{\utility}_\ast(o_t|\g{ao}_{<t}a_t) \\
        + \sum_{o_t} \agentc(o_t|\g{ao}_{<t}a_t) \log Z^{\alpha}(\g{ao}_{\leq t})
    \Bigg\}
\end{multline*}
where $Z^{\alpha}(\g{ao}_{\leq t})$ are the normalizing constants obtained for
the subsequent time step. This way the optimal action probabilities can be
computed recursively.

This result allows to recover the maximum expected utility solution, and more
specifically, the dynamic programming solution. Identify the value function as
$V^\alpha(\g{ao}_{<t}) \define \log Z^\alpha(\g{ao}_{<t})$, and the
instantaneous rewards as $r(a_t|\g{ao}_{<t}) \define
\ext{\utility}_\ast(a_t|\g{ao}_{<t})$ and $r(o_t|\g{ao}_{<t}a_t) \define
\ext{\utility}_\ast(o_t|\g{ao}_{<t}a_t)$. If one takes the limit $\alpha
\rightarrow 0$, then $\agentc(a_t|\g{ao}_{<t}) \rightarrow
\delta_{a^\ast}(a_t)$, where
\[
    a^\ast \define \max_{a_t} \Bigg\{
        r(a_t|\g{ao}_{<t})
        + \sum_{o_t} \agentc(o_t|\g{ao}_{<t}a_t)
            \Bigl[ r(o_t|\g{ao}_{<t}a_t) + V^0(\g{ao}_{\leq t}) \Bigr]
    \Bigg\}
\]
and where the value $V^0(\g{ao}_{<t})$ turns out to be given by the recursive
formula
\begin{align*}
    V^0(\g{ao}_{\leq T}) &\define 0, \\
    V^0(\g{ao}_{<t}) &= \max_{a_t} \Bigg\{
        r(a_t|\g{ao}_{<t})
        + \sum_{o_t} \agentc(o_t|\g{ao}_{<t}a_t)
            \Bigl[ r(o_t|\g{ao}_{<t}a_t) + V^0(\g{ao}_{\leq t}) \Bigr]
        \Bigg\}.
\end{align*}

Taking the limit $\alpha \rightarrow \infty$ puts all the emphasis of the
variational problem on the resource costs. This case yields
\[
    \agentc(a_t|\g{ao}_{<t}) = \ragentc(a_t|\g{ao}_{<t})
\]
as expected.

\subsection{Adaptive estimation}\label{sec:adaptive-estimation}

In an adaptive estimation problem one is confronted with an unknown symbol
source $\ragentc(o_t|\theta, o_{<t}) = \env(o_t|\theta, \hat{o}_{<t})$ indexed
by $\theta \in \Theta$ and chosen randomly as $\ragentc(\theta) = \env(\theta)$.
For this observation problem we can disregard the action variables and set
$\ext{\utility}_\ast = 0$. The probability measures $\gend$ and $\refd$ are
given by
\begin{align*}
    \gend(\theta) &= \ragentc(\theta),
    & \gend(o_t| \theta, o_{<t}) &= \ragentc(o_t|o_{<t}),
    \\
    \refd(\theta) &= \probc(\theta),
    & \refd(o_t| \theta, o_{<t}) &= \probc(o_t| o_{<t}).
\end{align*}
Replacing these distributions into~(\ref{eq:generalized-vp}) yields
\[
    -\alpha \sum_{\theta, o_{\leq T}} \ragentc(\theta)
        \prod_{t=1}^T \ragentc(o_t|\theta, o_{<t})
        \Bigg[
            \log \frac{\ragentc(\theta)}{\probc(\theta)}
            + \sum_{t=1}^T \log
                \frac{\ragentc(o_t|\theta, o_{<t})}
                     {\probc(o_t|o_{<t})}
        \Bigg]
\]
For the parameter $\theta$, we see that
\[
    \agentc(\theta) = \ragentc(\theta),
\]
and that the $t$-th summand of the functional can then be written as
\[
    -\alpha \sum_{\theta, o_{\leq T}} \ragentc(\theta) \prod_{t=1}^{t-1}
        \ragentc(o_t|\theta, o_{<t})
        \Bigg[
            \sum_{o_t} \ragentc(o_t|\theta, o_{<t})
                \log \frac{\ragentc(o_t|\theta, o_{<t})}
                          {\probc(o_t|o_{<t})}
        \Bigg]
\]
The solution to this variational problem is well-known in the literature
\citep{Opper1997,Opper1998} and is solved by the predictive distribution
\[
    \agentc(o_t|o_{<t})
    = \sum_{\theta} \ragentc(\theta|o_{<t}) \ragentc(o_t|\theta, o_{<t}),
\]
where the posterior $\ragentc(\theta|o_{<t})$ is computed according to Bayes'
rule.

\subsection{Adaptive control}\label{sec:adaptive-control}

In adaptive control problems the environment is not known a priori, but known
to belong to a set of possible environments $\ragentc(o_t|\theta,
\g{\hat{a}o}_{<t}\hat{a}_t) = \env(o_t|\theta, \g{a\hat{o}}_{<t}a_t)$
indexed by $\theta \in \Theta$ and
chosen randomly as $\ragentc(\theta) \define \env(\theta)$. We have also seen
that quantities that are estimated require the solution of a variational
problem that is local in time---in contrast to quantities that are controlled,
which require the solution of a variational problem that stretches over the
whole future. Can we devise an adaptive controller that is based on pure
estimation?

If we also happen to know a set of controllers $\ragentc(a_t|\theta,
\g{\hat{a}o}_{<t})$ for each of these environments (for instance, constructed
previously by solving the individual optimal control problems), then a Bayesian
rule for control can be devised---compare \cite{OrtegaBraun2010b}. Since in
pure estimation problems constraint utilities do not matter, we impose
$\ext{\utility}_\ast = 0$ for the sake of simplicity. The probability measures
$\gend$ and $\refd$ are given by
\begin{align*}
    \gend(\theta) &= \ragentc(\theta),
    & \gend(a_t|\theta, \g{ao}_{<t}) &= \ragentc(a_t|\theta, \g{\hat{a}o}_{<t}),
    & \gend(o_t|\theta, \g{ao}_{<t}a_t) &= \ragentc(o_t|\theta, \g{\hat{a}o}_{<t}\hat{a}_t),
    \\
    \refd(\theta) &= \probc(\theta),
    & \refd(a_t|\theta, \g{ao}_{<t}) &= \probc(a_t|\g{\hat{a}o}_{<t}),
    & \refd(o_t|\theta, \g{ao}_{<t}a_t) &= \probc(o_t|\g{\hat{a}o}_{<t}\hat{a}_t).
\end{align*}
Inserting them into~(\ref{eq:generalized-vp}) yields
\begin{small}
\[
    -\alpha \sum_{\theta,\g{ao}_{\leq T}}
        \ragentc(\theta) \biggl( \prod_{t=1}^T
            \ragentc(a_t|\theta, \g{\hat{a}o}_{<t})
            \ragentc(o_t|\theta, \g{\hat{a}o}_{<t}\hat{a}_t)
            \biggr)
            \sum_{t=1}^T \Bigg[
                \log\frac{\ragentc(a_t|\theta, \g{\hat{a}o}_{<t})}
                         {\probc(a_t|\g{\hat{a}o}_{<t})}
                +\log\frac{\ragentc(o_t|\theta, \g{\hat{a}o}_{<t}a_t)}
                          {\probc(o_t|\g{\hat{a}o}_{<t}a_t)}
            \Bigg]
\]
\end{small}
Again, for the parameter $\theta$, we see that
\[
    \agentc(\theta) = \ragentc(\theta).
\]
For the variational problem of the observation at time $t$ we can again
disregard the resource costs of the actions. Analogous to the solution for
adaptive estimation, the variational problem is equivalent to
\begin{small}
\begin{multline*}
    -\alpha
    \sum_{\theta,\g{ao}_{<t} a_t}
        \ragentc(\theta)
        \biggl( \prod_{\tau=1}^{t-1}
            \ragentc(a_\tau|\theta,\g{\hat{a}o}_{<\tau})
            \ragentc(o_\tau|\theta,\g{\hat{a}o}_{<\tau}\hat{a}_\tau)
        \biggr)
        \ragentc(a_t|\theta, \g{\hat{a}o}_{<t})
        \\ \times \sum_{o_t}
            \ragentc(o_t|\theta, \g{\hat{a}o}_{<t}\hat{a}_t)
            \log \frac{\ragentc(o_t|\theta, \g{\hat{a}o}_{<t}\hat{a}_t)}
                      {\probc(o_t|\theta, \g{\hat{a}o}_{<t}\hat{a}_t)}
\end{multline*}
\end{small}
which is solved by the predictive distribution
\[
    \agentc(o_t|\g{\hat{a}o}_{<t}\hat{a}_t)
    = \sum_{\theta} \ragentc(\theta|\g{\hat{a}o}_{<t}\hat{a}_t)
        \ragentc(o_t|\theta, \g{\hat{a}o}_{<t}\hat{a}_t).
\]
For actions, the procedure is identical. Thus, the variational problem for the
$t$-th action is given by
\[
    -\alpha
    \sum_{\theta,\g{ao}_{<t}}
        \ragentc(\theta)
        \biggl( \prod_{\tau=1}^{t-1}
            \ragentc(a_\tau|\theta,\g{\hat{a}o}_{<\tau})
            \ragentc(o_\tau|\theta,\g{\hat{a}o}_{<\tau}\hat{a}_\tau)
        \biggr)
        \sum_{a_t} \ragentc(a_t|\theta, \g{\hat{a}o}_{<t})
            \log \frac{\ragentc(a_t|\theta, \g{\hat{a}o}_{<t})}
                      {\probc(a_t|\theta, \g{\hat{a}o}_{<t})}
\]
again solved by the predictive distribution
\[
    \agentc(a_t|\g{\hat{a}o}_{<t})
    = \sum_{\theta} \ragentc(\theta|\g{\hat{a}o}_{<t})
        \ragentc(a_t|\theta, \g{\hat{a}o}_{<t}).
\]
This result has been previously reported as the \emph{Bayesian control rule}
\citep{OrtegaBraun2008, BraunOrtega2010}. By sampling from the
predictive distribution $\agentc(a_{t+1}|\g{\hat{a}o}_{\leq t})$ the agent can
solve adaptive control problems, such as bandit problems, adaptive linear
quadratic control problems and Markov decision problems with unknown transition
matrices.

\section{Discussion}

In this study we have used causal models to construct probability distributions representing I/O systems. As I/O systems both process input symbols and generate output symbols, their characterization requires both evidential and generative probabilities. The evidential probabilities (``plausibilities'' in the subjectivist sense of probability) allow the I/O system to infer properties about the input stream, while the generative probabilities (``propensities'' in the frequentist sense) prescribe the law to generate its output stream. The importance of distinguishing between input and output, more commonly known as the difference between \emph{seeing and doing}, and their impact on inference, lies at the heart of statistical causality \citep{Pearl2000, Spirtes2001}.

Based on the equivalence of information and utility, we have devised a variational principle to construct I/O systems. Structural similarities between utilities and information have been previously reported in the literature \citep{Candeal2001}. For the case of known environments, a duality between optimal control and estimation has been previously reported by \citet{Todorov2008}, where an exponential transformation mediates between the cost-to-go function and a probability distribution that acts as a backwards filter. For the case of optimally learning systems in unknown environments, a duality between utility and information has been reported by \citet{Belavkin2008}, considering the problem of optimal learning as a variational problem of expected utility maximization with dynamical information constraints. An information-theoretic approach to interactive learning based on principles from statistical physics has also been proposed by \citet{Still2009}. The use of the Kullback-Leibler divergence to measure deviations from a reference distribution as a cost function for control has been previously proposed by \citet{Todorov2006, Todorov2009} and by \citet{Kappen2009}. In these studies, transition probabilities of Markov systems were manipulated directly and the cost measured as a probabilistic deviation with respect to the passive dynamics of the system. Adaptive controllers based on the minimum relative entropy principle have been previously reported in \citet{OrtegaBraun2008} and in \citet{BraunOrtega2010}. The contribution of our study is to devise a single axiomatic framework that allows for the solution of both control and adaptation problems based on the equivalence of utility and information. This axiomatic framework leads to a single variational principle to solve both problems. The resulting controllers optimize a trade-off between maximization of a target utility function and resource costs and can hence be interpreted as bounded-rational actors.

The idea of bounded rationality through the consideration of information costs has been first proposed by \citet{Simon1982}. In game theory, information theory has been proposed to formalize bounded rational players whose degree of rationality is given by a temperature parameter trading off entropy and payoff \citep{Wolpert2004}. The distinction between disclosed and undisclosed information has also been studied extensively in the literature on game theory regarding problems of \emph{incomplete} or \emph{imperfect} information (see \citealt{Gibbons1992}, and \citealt{Osborne1999}). Like these previous studies, our work has obvious connections to information theory \citep{Shannon1948}, thermodynamics (see e.g.\  \citealt{Callen1985}) and statistical inference (see e.g.\  the maximum entropy principles in \citealt{Jaynes2003}).

\section{Conclusions}

The main contribution of the current paper is to derive axiomatically a
framework for bounded rationality. We propose to formalize agents as
probability distributions over I/O streams. Based on the idea that a free
system produces an outcome with higher probability if and only if it is more
desirable, we postulate three simple axioms relating utilities and
probabilities. We show that these axioms enforce a unique conversion law
between utility and probability (and thereby, information). Moreover, we show
that this relation can be characterized as a variational principle: given a
utility function, its conjugate probability measure maximizes the free utility
functional. We exhibit how constrained transformations of probability measures
can be characterized as a change in free utility and use this to formulate a
model of bounded rationality. Accordingly, one obtains a variational principle
to choose a probability measure that trades off the maximization of a target
utility function and the cost of the deviation from a reference distribution.
We show that optimal control, adaptive estimation and adaptive control problems
can be solved this way in a resource-efficient way. When resource costs are
ignored, the MEU principle is recovered. Our formalization might thus provide a
principled approach to bounded rationality that establishes a link to
information theory.

% Acknowledgements
%\acks{ACKS}

% Appendix

\vskip 0.2in
%GATHER{bibliography.bib}
\bibliography{bibliography}

\end{document}